\documentclass[conference]{IEEEtran}

\usepackage{amsmath,amssymb,amsfonts}
\usepackage{algorithm}
\usepackage{algorithmic}
\usepackage{booktabs}
\usepackage{multirow}
\usepackage{array}
\usepackage{graphicx}
\usepackage{xcolor}
\usepackage{cite}
\usepackage{url}
\usepackage{colortbl}\usepackage[table]{xcolor}
\usepackage{footnote}
\usepackage{enumerate}
\usepackage{placeins}
\usepackage{tikz}
\usetikzlibrary{arrows.meta, positioning, calc}
\begin{document}

\title{DE-2LS: Differential Evolution with Lightweight Late Local Search for Constrained Numerical Optimization}

\author{\IEEEauthorblockN{Dikshit Chauhan and Anupam Trivedi}
	\IEEEauthorblockA{Department of Electrical and Computer Engineering, \\National University of Singapore\\
		Email: dikshitchauhan608@gmail.com}}

\maketitle

\begin{abstract}
Constrained single-objective numerical optimization requires a careful balance among feasibility, objective convergence, and computational efficiency under a fixed function-evaluation budget. This paper proposes DE-2LS, a late-stage, locally search-enhanced variant of differential evolution built on the RDEx framework. The proposed method preserves the original RDEx components, including mutation and crossover operators, success-history adaptation, archive mechanism, population-size reduction, and $\epsilon$-based constraint handling. A lightweight coordinate-pattern local search is added as a guarded polishing component around the current best solution. It is activated only in the late stage of the run, uses a small evaluation budget, and accepts candidates through a feasibility-aware comparison rule. Ablation results show that the finalized DE-2LS configuration achieves the best U-score among all tested variants, confirming that controlled late-stage refinement is more effective than aggressive or premature local search. In the direct comparison with RDEx, DE-2LS achieves a 5.58\% gain in U-score. In the four-algorithm comparison, DE-2LS obtains the highest overall U-score of 80968 and the best total rank of 48 among RDEx, CL-SRDE, and UDE-III. These results indicate that DE-2LS improves the exploitation capability of the RDEx-based search framework while preserving its speed advantage under the combined speed-accuracy scoring criterion. The source code of DE-2LS is available at \url{https://github.com/ChauhanDikshit?tab=repositories}.
\end{abstract}


\section{Introduction}
\IEEEPARstart{D}{ifferential} evolution (DE) is one of the most widely used population-based algorithms for continuous black-box optimization. Since its introduction by Storn and Price \cite{Storn1997DE}, DE has attracted continued attention due to its simple mutation-crossover-selection cycle, a small number of control parameters, and strong empirical performance on multimodal and nonseparable search landscapes \cite{chauhan2025advancements}. Over the last decade, the most competitive DE variants have moved beyond fixed control parameters and single mutation rules, incorporating success-history adaptation, ensemble mutation strategies, archive-assisted search, population-size reduction, restart strategies, covariance information, and local-search components \cite{Tanabe2014LSHADE,Brest2017jSO,Kumar2017EBOCMAR,Sallam2020IMODE,Biedrzycki2022NLSHADERSP}.

The IEEE CEC numerical optimization competitions have played an important role in shaping these developments. In bound-constrained settings, algorithms such as EBOwithCMAR, jSO, IMODE, NL-SHADE-RSP, EA4Eig, L-SRTDE, and RDEx have represented different ways to improve the balance between exploration and exploitation. In constrained optimization, the problem becomes more difficult because the algorithm must optimize the objective value while also satisfying inequality and equality constraints. Therefore, successful algorithms usually combine a strong search engine with a constraint-handling mechanism such as Deb's feasibility rule, $\epsilon$-constrained selection, penalty-based ranking, or hybrid feasibility-aware comparison \cite{Wu2017CEC2017COP,Fan2018LSHADE44IEpsilon,Trivedi2017UDE,UDEIII2024,Tao2026RDExCSOP}.

Table~\ref{tab:cec_de_history} summarizes representative DE-related methods and competition winners/top performers from 2017 to 2025. The purpose of the table is not to compare exact scores across different benchmark tracks, since the competition settings changed across years, but to highlight the algorithmic trend. Early winners and top DE variants relied on success-history adaptation and hybridization with covariance or local search ideas. More recent algorithms increasingly emphasize multiple search operators, ranking-based mutation, restart or stagnation control, and evaluation-budget-aware scoring.

\begin{table*}[!t]
\centering
\caption{Representative CEC competition winners, top-performing DE variants, and related algorithmic mechanisms from 2017-2025.}
\label{tab:cec_de_history}
\renewcommand{\arraystretch}{1.15}
\begin{tabular}{p{0.08\textwidth}p{0.17\textwidth}p{0.23\textwidth}p{0.34\textwidth}p{0.10\textwidth}}
\toprule
Year & Representative algorithm & Track/context & Main mechanisms & Ref. \\
\midrule
2017 & EBOwithCMAR & CEC 2017 bound-constrained single-objective competition winner/top method & Butterfly optimizer enhanced with covariance-matrix-adapted retreat phase for stronger local search & \cite{Kumar2017EBOCMAR} \\
2017 & jSO & CEC 2017 single-objective bound-constrained DE variant & Success-history adaptation, external archive, current-to-pbest mutation, population-size reduction & \cite{Brest2017jSO} \\
2017 & LSHADE44-IEpsilon/UDE & CEC 2017 constrained optimization top DE variants & Improved $\epsilon$ constraint handling, multi-operator DE, ranking-based mutation & \cite{Fan2018LSHADE44IEpsilon,Trivedi2017UDE} \\
2018 & LSHADE-SPACMA & CEC-style bound-constrained optimization & Hybrid L-SHADE and CMA-ES style adaptation for nonseparable landscapes & \cite{Mohamed2017LSHADESPACMA} \\
2019 & DISH/jDE & CEC 2019 100-digit challenge context & Long-budget DE variants, enhanced mutation/crossover, archive and parameter adaptation & \cite{Viktorin2020DISHXX} \\
2020 & IMODE & CEC 2020 single-objective bound-constrained competition winner/top method & Improved multi-operator DE, adaptive operator selection, population-size control & \cite{Sallam2020IMODE} \\
2021 & NL-SHADE-RSP family & CEC 2021 bound-constrained competition winner/top method & Nonlinear population-size reduction, rank-based selective pressure, SHADE-style memory & \cite{Biedrzycki2022NLSHADERSP} \\
2022 & EA4Eig & CEC 2022 bound-constrained winner & Ensemble optimizer with eigen-coordinate crossover and multiple components & \cite{Biedrzycki2024RankingCEC2022} \\
2024 & L-SRTDE/mLSHADE-RL & CEC 2024 bound-constrained single-objective optimization & Rank-based DE, restart, multi-operator ensemble, late local search & \cite{stanovov2024success,mLSHADERL2024} \\
2024 & UDE-III & CEC 2024 constrained real-parameter optimization & Unified DE, subpopulation design, ranking mutation, enhanced $\epsilon$ constraint handling, stagnation control & \cite{UDEIII2024} \\
2025 & RDEx-CSOP & CEC 2025 numerical optimization tracks & Reconstructed DE, exploitation-biased branch, feasibility-aware ranking, U-score-oriented performance & \cite{Tao2026RDExCSOP} \\
\bottomrule
\end{tabular}
\end{table*}

RDEx \cite{Tao2026RDExCSOP} provides a strong baseline for constrained CEC-style optimization by combining success-history parameter adaptation, feasibility-aware comparison, an exploitation-biased branch, and a decreasing population structure. In the pairwise U-score setting considered in this work, final solution quality and relative ranking against other algorithms are important. Therefore, even a small improvement in the final feasible objective value can improve the overall score. The motivation of this work is to enhance the final-stage exploitation capability of RDEx while preserving its main search behavior.

Replacing RDEx operators or constraint handling may compromise the properties that make RDEx strong. Therefore, the proposed approach is deliberately conservative: the original RDEx implementation is kept unchanged, and a lightweight local search (LS) is added only as a late-stage polishing mechanism. This design follows a ``do no harm'' philosophy: RDEx performs the global and feasibility-oriented search, while LS is allowed to refine the best solution only after the search has reached a mature stage. The main contributions of this paper are as follows:
\begin{enumerate}[(i)]
    \item A late-stage local-search-enhanced DE variant, named DE-2LS, is proposed for constrained single-objective numerical optimization.


    \item A lightweight coordinate-pattern local-search module is introduced around the current best solution to improve late-stage exploitation without replacing the population-based search process.

    \item A feasibility-aware acceptance rule is used in the local search, ensuring that a candidate is accepted only if it improves the incumbent under the same constrained-optimization logic used by RDEx.

     \item The local search is activated only in the late stage with a small evaluation budget, preventing premature or excessive exploitation.

    \item Ablation and comparative results show that DE-2LS improves the original RDEx and achieves the best overall U-score among the compared algorithms.
\end{enumerate}

The remainder of this paper is organized as follows. Section~\ref{sec:de} reviews the basic DE framework. Section~\ref{sec:rdex} summarizes RDEx and discusses its potential weakness. Section~\ref{sec:proposed} presents the proposed DE-2LS algorithm. Section~\ref{sec:results} reports the experimental setting and discusses the preliminary pairwise U-score results. Section~\ref{sec:conclusion} concludes the paper.

\section{Differential Evolution}
\label{sec:de}
DE is a population-based evolutionary algorithm for real-parameter optimization \cite{Storn1997DE}. Consider the minimization problem
\begin{equation}
    \min_{\mathbf{x}\in\Omega} f(\mathbf{x}),
\end{equation}
where $\mathbf{x}=[x_1,x_2,\ldots,x_D]^T$ is a $D$-dimensional decision vector and $\Omega$ denotes the search domain. At generation $g$, DE maintains a population $\{\mathbf{x}_{i,g}\}_{i=1}^{N_g}$. For each target vector $\mathbf{x}_{i,g}$, a mutant vector is generated by combining other population members. A classical mutation strategy is
\begin{equation}
    \mathbf{v}_{i,g}=\mathbf{x}_{r_1,g}+F(\mathbf{x}_{r_2,g}-\mathbf{x}_{r_3,g}),
\end{equation}
where $r_1$, $r_2$, and $r_3$ are mutually distinct indices and $F$ is the scaling factor. The binomial crossover operator generates a trial vector $\mathbf{u}_{i,g}$ as
\begin{equation}
    u_{j,i,g}=\begin{cases}
    v_{j,i,g}, & \text{if } rand_j \leq CR \text{ or } j=j_{rand},\\
    x_{j,i,g}, & \text{otherwise},
    \end{cases}
\end{equation}
where $CR$ is the crossover rate and $j_{rand}$ guarantees that at least one component comes from the mutant vector. For unconstrained minimization, selection is usually performed as
\begin{equation}
    \mathbf{x}_{i,g+1}=\begin{cases}
    \mathbf{u}_{i,g}, & \text{if } f(\mathbf{u}_{i,g})\leq f(\mathbf{x}_{i,g}),\\
    \mathbf{x}_{i,g}, & \text{otherwise}.
    \end{cases}
\end{equation}

For constrained optimization, the problem is formulated as
\begin{equation}
\begin{aligned}
    \min_{\mathbf{x}} \quad & f(\mathbf{x})\\
    \text{s.t.}\quad & g_i(\mathbf{x})\leq 0, \quad i=1,\ldots,m,\\
    & h_j(\mathbf{x})=0, \quad j=1,\ldots,q,\\
    & L_k\leq x_k\leq U_k, \quad k=1,\ldots,D.
\end{aligned}
\end{equation}
The total constraint violation (CV) can be written as
\begin{equation}
    CV(\mathbf{x})=\sum_{i=1}^{m}\max\{0,g_i(\mathbf{x})\}+\sum_{j=1}^{q}\max\{0,|h_j(\mathbf{x})|-\delta\},
\end{equation}
where $\delta$ is a tolerance for equality constraints. A feasibility-aware comparison is then required. A typical $\epsilon-$based rule treats a solution as quasi-feasible if $CV(\mathbf{x})\leq\epsilon$ and compares objective values inside the quasi-feasible region while comparing constraint violations outside it.

\section{Base RDEx Framework}
\label{sec:rdex}
RDEx is a reconstructed DE framework designed for fixed-budget CEC-style numerical optimization \cite{Tao2026RDExCSOP}. In the constrained single-objective setting, RDEx combines several mechanisms that are important for robust performance.

First, RDEx uses success-history parameter adaptation. Similar to SHADE/L-SHADE-type algorithms, successful values of $F$ and $CR$ are stored and reused to guide future parameter sampling. This allows the algorithm to adapt its search step size and crossover behavior based on the observed success of previous generations.

Second, RDEx uses feasibility-aware ranking and $\epsilon$-constraint handling. This is essential for constrained CEC problems, where an infeasible solution with a very low objective value should not dominate a feasible or nearly feasible solution. The constraint-handling rule allows the algorithm to balance objective minimization and feasibility improvement throughout the run.

Third, RDEx incorporates an exploitation-biased search branch. This component increases selective pressure toward better solutions and supports fast convergence, which is useful under U-score-based evaluation, where both speed and final quality can influence the score.

Fourth, RDEx applies population-size reduction. The algorithm starts with a relatively larger population to maintain diversity and gradually reduces the active population to increase exploitation in the later stage. This strategy has been widely used in strong DE variants because it supports exploration early and convergence later \cite{Tanabe2014LSHADE,Brest2017jSO}.

\subsection{Potential Weakness of RDEx}
Although RDEx is highly competitive, its final-stage behavior can still be limited by the population-level evolutionary operators. After a feasible basin has been found, the remaining search budget may be spent generating trial vectors that are too coarse or too stochastic to efficiently refine the incumbent solution. This issue is especially relevant in the pairwise scoring setting, where the final relative quality against other algorithms can be decisive.

The weakness can be summarized as follows:
\begin{itemize}
    \item RDEx is strong in global search and feasibility discovery, but its final local exploitation may be insufficient on some functions.
    \item Late-stage population reduction increases exploitation, but it does not guarantee fine coordinate-level improvement around the best solution.
    \item Aggressively changing RDEx operators may harm its established feasibility behavior and speed advantage.
\end{itemize}
These observations motivate adding a small and controlled local-search module only in the late stage.

\section{Proposed DE-2LS Algorithm}
\label{sec:proposed}
This section presents DE-2LS, a late-stage local-search enhanced version of
RDEx. The proposed method preserves the original RDEx search framework and adds
a lightweight local-search module only as a final polishing mechanism. The main
motivation is to improve the final solution quality without weakening the fast feasibility-oriented search behavior of RDEx.

\subsection{Design Principle}
\label{subsec:design_principle}
DE-2LS is designed under a conservative ``do-no-harm'' principle. The original
RDEx components are retained, including the mutation and crossover operators,
success-history memory update, archive management, $\epsilon$-based constraint
handling, and population-size reduction. The additional component is a coordinate pattern
local search applied to the current best solution during the late stage of the
search.

This design is motivated by the observation that RDEx is effective in global
search and feasibility discovery, but its final-stage exploitation can be further
improved. Therefore, instead of modifying the main evolutionary engine, DE-2LS
adds a local refinement step after most of the evolutionary search has
already been performed.

\subsection{Why Late Local Search?}
Inspired by the local refinement idea used in EBOwithCMAR~\cite{Kumar2017EBOCMAR}, the local search can improve solution accuracy, but using it too early may be harmful. In the early and middle stages, the population is still exploring the landscape and discovering feasible regions. If local search is activated too early, it can consume function evaluations near a solution that is not yet in a promising basin. This can reduce population diversity and slow down feasibility discovery.

DE-2LS therefore uses LS only in the late stage. The selected setting permits LS after \texttt{LS\_SR} of the maximum function evaluations and forces a very late call after \texttt{LS\_FSR} if LS has not yet been used. This gives RDEx most of the budget to perform its original evolutionary search, while reserving a small portion for final exploitation.

\subsection{Late Local Search Mechanism}
\label{subsec:ls_mechanism}

The local-search component is activated only in the late stage of the run. Let
$\mathbf{x}^{b}$ denote the current best solution. DE-2LS performs a
coordinate-pattern search around $\mathbf{x}^{b}$. The initial coordinate step is
computed as
\begin{equation}
    \Delta_0 = \texttt{LS\_ISR}\cdot (U_f-L_f),
\end{equation}
where \texttt{LS\_ISR} is the initial step-rate parameter, and $L_f$ and $U_f$ are the
function-specific lower and upper bounds. For each coordinate $d$, two candidate
moves are examined:
\begin{equation}
    x_d^{+}=x_d^{b}+\Delta, \qquad
    x_d^{-}=x_d^{b}-\Delta .
\end{equation}
If a candidate violates the search bounds, it is repaired using the same
bound-handling philosophy as RDEx. A first-improvement rule is adopted: once a
feasibility-aware better point is found, the local-search center is immediately
updated. If no improvement is found after scanning all coordinates, the step size
is reduced as
\begin{equation}
    \Delta \leftarrow \frac{\Delta}{2}.
\end{equation}
The local search terminates when its evaluation budget is exhausted, the global maximum function-evaluation budget is reached, or the step size becomes smaller
than the minimum step threshold.

\subsection{Feasibility-Aware Acceptance}
\label{subsec:ls_acceptance}
For a candidate solution $\mathbf{x}$, DE-2LS computes both the objective value $f(\mathbf{x})$ and total constraint violation $CV(\mathbf{x})$. An LS candidate $\mathbf{x}^{new}$ is considered better than the incumbent $\mathbf{x}^{old}$ if it satisfies the feasibility-aware rule:
\begin{align}
&\mathbf{x}^{new} \prec \mathbf{x}^{old}
\Longleftrightarrow\nonumber\\
&\begin{cases}
f(\mathbf{x}^{new}) < f(\mathbf{x}^{old}), 
& CV(\mathbf{x}^{new}) \leq \epsilon,\ CV(\mathbf{x}^{old}) \leq \epsilon,\\
CV(\mathbf{x}^{new}) \leq \epsilon, 
& CV(\mathbf{x}^{old}) > \epsilon,\\
CV(\mathbf{x}^{new}) < CV(\mathbf{x}^{old}), 
& CV(\mathbf{x}^{new}) > \epsilon,\ CV(\mathbf{x}^{old}) > \epsilon.
\end{cases}
\end{align}
This rule prevents LS from accepting a low-objective but highly infeasible solution. It also allows LS to reduce constraint violation when the current best solution is still infeasible.

\subsection{Triggering Logic}
The LS module is not called at every generation. Instead, it is triggered only when one of the following conditions is satisfied:
\begin{itemize}
    \item the search progress is at least \texttt{LS\_SR}, the global best has stagnated for at least eight generations, and the incumbent is feasible or nearly feasible; or
    \item the search progress is at least \texttt{LS\_FSR}, in which case a final polishing call is allowed even if the feasible-or-near condition is not strongly satisfied.
\end{itemize}
This logic is designed to avoid premature exploitation. The first condition supports polishing only when RDEx appears to have stopped improving. The second condition ensures that a final LS opportunity is not missed near the end of the run.

\subsection{Population Injection After LS}
If LS improves the global best solution, the improved solution is injected back into the RDEx population. Importantly, DE-2LS does not replace the whole population and does not reset the success-history memory. Only the improved incumbent is inserted into a suitable population position so that subsequent RDEx operations can use the refined solution. This conservative injection protects RDEx's adaptive search dynamics. Algorithm~\ref{alg:rdexls} summarizes the proposed DE-2LS framework.

\begin{algorithm}[!h]
\caption{DE-2LS: Late-Stage Local Search Enhanced RDEx}
\label{alg:rdexls}
\begin{algorithmic}[1]
\STATE Initialize RDEx population, fitness, constraint violation, memories, archive, and control parameters.
\WHILE{function-evaluation budget is not exhausted}
    \STATE Generate RDEx trial vectors using the original RDEx operators.
    \STATE Evaluate objective value and constraint violation of trial vectors.
    \STATE Apply RDEx feasibility-aware selection and update the population.
    \STATE Update RDEx success-history memories, archive, and population size.
    \STATE Update the global best solution and its stagnation counter.
    \IF{progress $\geq \texttt{LS\_SR}$ and stagnation $\geq \texttt{LS\_SG}$ and best is feasible/nearly feasible}
        \STATE Run coordinate-pattern LS around the current best solution.
    \ELSIF{progress $\geq \texttt{LS\_FSR}$ and LS call limit is not reached}
        \STATE Run one very-late coordinate-pattern LS call.
    \ENDIF
    \IF{LS returns a feasibility-aware better solution}
        \STATE Update global best and inject improved solution into the population.
    \ENDIF
\ENDWHILE
\STATE Return the final best solution and recorded progress curve.
\end{algorithmic}
\end{algorithm}

\section{Results and Discussion}
\label{sec:results}

\subsection{Experimental Setting}
\label{subsec:experimental_setting}
The proposed DE-2LS is evaluated on the 28 constrained single-objective
real-parameter optimization problems from the CEC constrained benchmark suite.
Following the competition setting, the dimension is fixed at $D=30$, and the
maximum number of function evaluations is $\mathrm{MaxFEs}=20000\times D$.
Each algorithm is executed for 25 independent runs on each function. The compared algorithms are DE-2LS, the original RDEx, CL-SRDE, and UDE-III. All algorithms use the same stopping criterion and evaluation budget. A larger U-score indicates better overall performance. When algorithm $a$ is compared against a baseline algorithm $b$, the relative gain is computed as
\begin{equation}
\text{Relative Gain}(a \mid b)
=
\frac{\mathcal{U}_a-\mathcal{U}_b}{\mathcal{U}_b}\times 100,
\label{eq:relative_gain}
\end{equation}
where $\mathcal{U}_a$ and $\mathcal{U}_b$ are the corresponding overall U-scores. A positive value indicates that algorithm $a$ outperforms the baseline.
\subsubsection{Procedure for U-score}
For the U-score calculation, we follow  the CEC 2026 constrained optimization evaluation protocol\footnote{\url{https://github.com/P-N-Suganthan/2026-CEC}}; each run stores the best-so-far value
\texttt{Min\_EV} and the lowest constraint violation \texttt{LCV} after
initialization and then every $10D$ evaluations until \texttt{MaxFEs}. Therefore, each run contains 2001 saved points. Let $E_{a,r}^{(f)}(t)$ and
$C_{a,r}^{(f)}(t)$ denote the saved \texttt{Min\_EV} and \texttt{LCV},
respectively, for algorithm $a$, run $r$, function $f$, and sampling point $t$.
The LCV is computed as:
\begin{equation}
    C_{a,r}^{(f)}(t)=\min_{1\leq i\leq NP} CV(P_i(t)),
\end{equation}
where $P_i(t)$ is the $i$th individual in the population at sampling point $t$.
If no feasible solution exists at a sampling point, \texttt{Min\_EV} is recorded
as \texttt{NaN}.

Let $m$ be the number of compared algorithms and $n$ be the number of independent
runs. For each function, all $mn$ trials are compared pairwise. The number of
trial pairs is
\begin{equation}
    N_{\mathrm{pair}}=\frac{mn(mn-1)}{2}.
\end{equation}
For two trials $x$ and $y$, let $E_x^{\star}$ and $C_x^{\star}$ denote the final
saved error and constraint violation at \texttt{MaxFEs}. A trial is feasible if
$C_x^{\star}\leq \epsilon_{\mathrm{CV}}$ and $E_x^{\star}$ is finite.

For pairwise scoring, define the smaller-is-better comparison function
\begin{equation}
\label{eq:rho}
\rho(u,v;\epsilon)=
\begin{cases}
1, & u < v-\epsilon,\\
0.5, & |u-v|\leq \epsilon,\\
0, & u > v+\epsilon.
\end{cases}
\end{equation}
The value $\rho(u,v;\epsilon)$ represents the point awarded to the first trial
when comparing $u$ and $v$.

The pairwise accuracy point awarded to trial $x$ against trial $y$ is computed
as
\begin{equation}
\label{eq:pairwise_accuracy}
A(x,y)=
\begin{cases}
\rho(E_x^{\star},E_y^{\star};\epsilon_{\mathrm{EV}}),
& x,y \text{ are feasible},\\
\rho(C_x^{\star},C_y^{\star};\epsilon_{\mathrm{CV}}),
& x,y \text{ are infeasible},\\
1, & x \text{ is feasible and } y \text{ is infeasible},\\
0, & x \text{ is infeasible and } y \text{ is feasible},
\end{cases}
\end{equation}
 Therefore, feasible trials are always
preferred over infeasible trials. If both trials are feasible, the final
\texttt{Min\_EV} values are compared. If both trials are infeasible, the final
\texttt{LCV} values are compared.

The accuracy score of algorithm $a$ on function $f$ is the sum of all pairwise
accuracy points obtained by its $n$ trials:
\begin{equation}
\label{eq:algorithm_accuracy}
    \mathcal{A}_{a}^{(f)}
    =
    \sum_{x\in \mathcal{T}_{a}^{(f)}}
    \sum_{\substack{y\in \mathcal{T}^{(f)}\\y\neq x}}
    A(x,y),
\end{equation}
where $\mathcal{T}_{a}^{(f)}$ is the set of trials of algorithm $a$ on function
$f$, and $\mathcal{T}^{(f)}$ is the set of all trials from all algorithms on
function $f$.

The speed score is computed using the first sampling point at which a trial
reaches a pairwise reference value. For two final feasible trials, the reference
value is the worse final error value:
\begin{equation}
\label{eq:theta_error}
    \theta_E(x,y)=\max(E_x^{\star},E_y^{\star}).
\end{equation}
The first reaching index of trial $x$ is then defined as
\begin{equation}
\label{eq:tau_error}
    \tau_E(x,y)=
    \min \left\{t \mid E_x(t)\leq \theta_E(x,y)+\epsilon_{\mathrm{EV}}\right\}.
\end{equation}
If the reference value is never reached, $\tau_E(x,y)$ is set to infinity.

If one or both trials are infeasible at the final sampling point, the comparison
is based on the lowest constraint violation. The pairwise reference value is
\begin{equation}
\label{eq:theta_cv}
    \theta_C(x,y)=\max(C_x^{\star},C_y^{\star}),
\end{equation}
and the first reaching index is
\begin{equation}
\label{eq:tau_cv}
    \tau_C(x,y)=
    \min \left\{t \mid C_x(t)\leq \theta_C(x,y)+\epsilon_{\mathrm{CV}}\right\}.
\end{equation}

The speed point awarded to trial $x$ against trial $y$ is therefore
\begin{equation}
\label{eq:pairwise_speed}
S(x,y)=
\begin{cases}
\rho(\tau_E(x,y),\tau_E(y,x);0),
& x,y \text{ are feasible},\\
\rho(\tau_C(x,y),\tau_C(y,x);0),
& \text{otherwise}.
\end{cases}
\end{equation}
Thus, the trial that reaches the pairwise reference value earlier receives one
point. If both trials reach it at the same sampling point, both trials receive
0.5 points.

The speed score of algorithm $a$ on function $f$ is computed as
\begin{equation}
\label{eq:algorithm_speed}
    \mathcal{S}_{a}^{(f)}
    =
    \sum_{x\in \mathcal{T}_{a}^{(f)}}
    \sum_{\substack{y\in \mathcal{T}^{(f)}\\y\neq x}}
    S(x,y).
\end{equation}

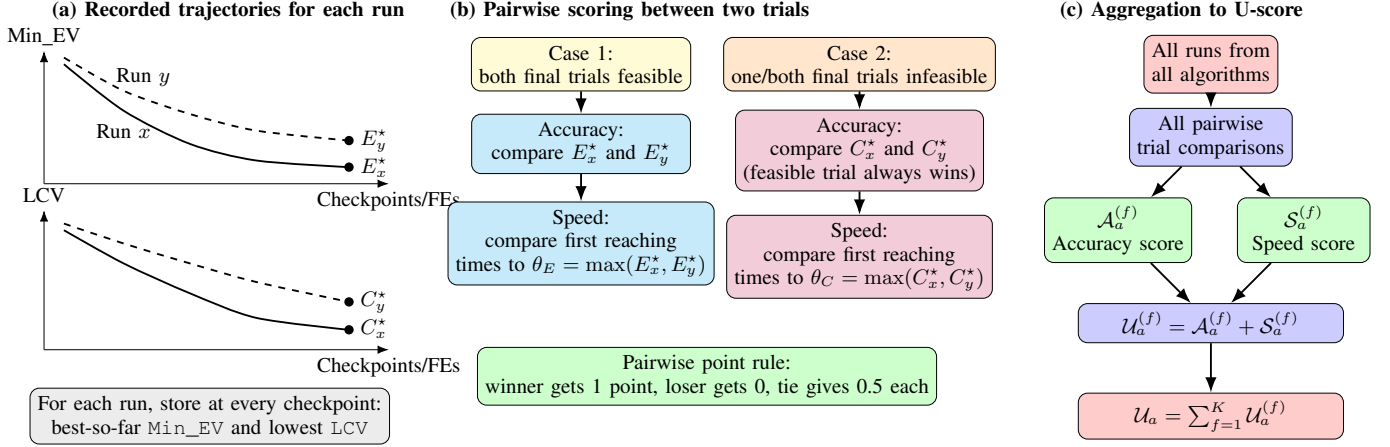
\begin{figure*}[!t]
\centering
\resizebox{1\linewidth}{!}{\begin{tikzpicture}[
    font=\small,
    >=Latex,
    box/.style={draw, rounded corners, align=center, minimum height=0.5cm, minimum width=2cm},
    smallbox/.style={draw, rounded corners, align=center, minimum height=0.7cm, minimum width=2.3cm},
    arr/.style={->, thick}
]

\node[anchor=west] at (0,4.8) {\textbf{(a) Recorded trajectories for each run}};

\draw[->] (0,2.2) -- (5.2,2.2) node[below] {Checkpoints/FEs};
\draw[->] (0,2.2) -- (0,4.2) node[above] {Min\_EV};

\draw[thick] plot[smooth] coordinates {(0.3,4.0) (1.2,3.3) (2.2,2.8) (3.2,2.55) (4.6,2.45)};
\draw[thick,dashed] plot[smooth] coordinates {(0.3,4.1) (1.2,3.6) (2.2,3.25) (3.2,3.0) (4.6,2.85)};
\fill (4.6,2.45) circle (2pt);
\fill (4.6,2.85) circle (2pt);

\node[right] at (4.65,2.45) {$E_x^\star$};
\node[right] at (4.65,2.85) {$E_y^\star$};
\node at (1.2,3.0) {Run $x$};
\node at (1.5,3.8) {Run $y$};

\draw[->] (0,-0.3) -- (5.2,-0.3) node[below] {Checkpoints/FEs};
\draw[->] (0,-0.3) -- (0,1.8) node[above] {LCV};

\draw[thick] plot[smooth] coordinates {(0.3,1.5) (1.2,1.0) (2.2,0.55) (3.2,0.18) (4.6,0.0)};
\draw[thick,dashed] plot[smooth] coordinates {(0.3,1.6) (1.2,1.3) (2.2,1.0) (3.2,0.72) (4.6,0.42)};
\fill (4.6,0.0) circle (2pt);
\fill (4.6,0.42) circle (2pt);

\node[right] at (4.65,0.02) {$C_x^\star$};
\node[right] at (4.65,0.42) {$C_y^\star$};

\node[box, fill=gray!15, minimum width=4.5cm] at (2.5,-1.3)
{For each run, store at every checkpoint:\\
best-so-far \texttt{Min\_EV} and lowest \texttt{LCV}};

\node[anchor=west] at (6.0,4.8) {\textbf{(b) Pairwise scoring between two trials}};

\node[smallbox, fill=yellow!20] (case1) at (8.1,4.0)
{Case 1:\\ both final trials feasible};

\node[smallbox, fill=orange!20] (case2) at (12.3,4.0)
{Case 2:\\ one/both final trials infeasible};

\node[box, fill=cyan!20, minimum width=3.3cm] (acc1) at (8.1,2.8)
{Accuracy:\\ compare $E_x^\star$ and $E_y^\star$};

\node[box, fill=cyan!20, minimum width=3.3cm] (spd1) at (8.1,1.3)
{Speed:\\ compare first reaching\\ times
to $\theta_E=\max(E_x^\star,E_y^\star)$};

\node[box, fill=purple!20, minimum width=3.5cm] (acc2) at (12.3,2.7)
{Accuracy:\\ compare $C_x^\star$ and $C_y^\star$\\
(feasible trial always wins)};

\node[box, fill=purple!20, minimum width=3.5cm] (spd2) at (12.3,1.1)
{Speed:\\ compare first reaching \\times
to $\theta_C=\max(C_x^\star,C_y^\star)$};

\draw[arr] (case1) -- (acc1);
\draw[arr] (acc1) -- (spd1);

\draw[arr] (case2) -- (acc2);
\draw[arr] (acc2) -- (spd2);

\node[box, fill=green!20, minimum width=6.4cm] at (10.0,-0.7)
{Pairwise point rule:\\
winner gets 1 point, loser gets 0, tie gives 0.5 each};

\node[anchor=west] at (15.2,4.8) {\textbf{(c) Aggregation to U-score}};

\node[box, fill=red!20]   (allruns) at (17.6,4.0) {All runs from\\all algorithms};
\node[box, fill=blue!20]  (pairs)   at (17.6,2.9) {All pairwise\\trial comparisons};
\node[box, fill=green!20] (A)       at (16.2,1.5) {$\mathcal{A}_a^{(f)}$\\Accuracy score};
\node[box, fill=green!20] (S)       at (19.0,1.5) {$\mathcal{S}_a^{(f)}$\\Speed score};
\node[box, fill=blue!20, minimum width=4.0cm] (Ufun) at (17.6,0.1)
{$\mathcal{U}_a^{(f)}=\mathcal{A}_a^{(f)}+\mathcal{S}_a^{(f)}$};
\node[box, fill=red!20, minimum width=4.0cm] (Utot) at (17.6,-1.3)
{$\mathcal{U}_a=\sum_{f=1}^{K}\mathcal{U}_a^{(f)}$};

\draw[arr] (allruns) -- (pairs);
\draw[arr] (pairs) -- (A);
\draw[arr] (pairs) -- (S);
\draw[arr] (A) -- (Ufun);
\draw[arr] (S) -- (Ufun);
\draw[arr] (Ufun) -- (Utot);

\end{tikzpicture}}
\caption{Illustration of the pairwise U-score procedure for CSOPs. 
(a) For each run, the best-so-far feasible objective value \texttt{Min\_EV} and the lowest constraint violation \texttt{LCV} are recorded at fixed checkpoints. 
(b) For a pair of trials, the accuracy and speed components are computed in a feasibility-aware manner. If both final trials are feasible, accuracy compares the final \texttt{Min\_EV} values and speed compares the first checkpoint at which each trial reaches the pairwise reference level $\theta_E=\max(E_x^\star,E_y^\star)$. If one or both final trials are infeasible, accuracy is based on the final \texttt{LCV} values (with a feasible trial always preferred), and speed compares the first checkpoint at which each trial reaches $\theta_C=\max(C_x^\star,C_y^\star)$. 
(c) Pairwise accuracy and speed points are accumulated over all trial pairs to obtain the function-wise score $\mathcal{U}_a^{(f)}$, which is then summed over all functions to produce the final U-score.}
\label{fig:uscore_csop_procedure}
\end{figure*}

The U-score of algorithm $a$ on function $f$ is the sum of its accuracy and speed
scores:
\begin{equation}
\label{eq:function_uscore}
    \mathcal{U}_{a}^{(f)}
    =
    \mathcal{A}_{a}^{(f)}+\mathcal{S}_{a}^{(f)}.
\end{equation}
The final U-score over all $K=28$ functions is calculated as
\begin{equation}
\label{eq:final_uscore}
    \mathcal{U}_{a}
    =
    \sum_{f=1}^{K}\mathcal{U}_{a}^{(f)}.
\end{equation}
For each function, algorithms are ranked according to $\mathcal{U}_{a}^{(f)}$ in
descending order. Rank 1 denotes the best algorithm. The total rank is obtained
by summing the function-wise ranks across all benchmark functions. The procedure of U-score calculation is illustrated in Fig. \ref{fig:uscore_csop_procedure}.
\subsection{Ablation Study of DE-2LS}
\label{subsec:ablation}
Table~\ref{tab:rdex_2ls_ablation} reports the ablation results of different
DE-2LS variants. In the table, BR01 and BR015 denote variants with a
larger local-search budget, 1\% and 1.5\%, respectively.  SR80
and SR90 denote variants with local search starting at 80\% and 90\% of the
evaluation budget, respectively. SR80SG12 denotes the variant with an earlier
starting rate and a stricter stagnation threshold. MC1 denotes a single local-search call. BR005MC3
denotes a smaller local search budget with three local search calls. ISR012 and ISR020 denote variants with different initial local search step sizes: 0.012 and 0.020, respectively. 

\begin{table}[!t]
\centering
\caption{Ablation comparison of DE-2LS variants.}
\label{tab:rdex_2ls_ablation}
\renewcommand{\arraystretch}{1.15}
\resizebox{\columnwidth}{!}{
\begin{tabular}{lrrrr}
\toprule
Variant name & Accuracy & Speed & U-score & Rank \\
\midrule
DE-2LS-BR01      & 87625.5 & 87549 & 175174.5 & 143 \\
DE-2LS-SR80       & 85551 & 86618.5 & 172169.5 & 169 \\
DE-2LS-SR80SG12    & 87270 & 84520.5 & 171790.5 & 169 \\
DE-2LS-SR90       & 88686.5 & 88495 & 177181.5 & \textbf{134} \\
\cellcolor{gray!25}\textbf{DE-2LS} 
                   & \cellcolor{gray!25}\textbf{88801.5} 
                   & \cellcolor{gray!25}\textbf{89947.5} 
                   & \cellcolor{gray!25}\textbf{178749} 
                   & \cellcolor{gray!25}\textbf{134} \\
DE-2LS-BR015      & 85999.5 & 86401 & 172400.5 & 168 \\
DE-2LS-MC1        & 86733.5 & 87179.5 & 173913 & 156 \\
DE-2LS-BR005MC3    & 87041.5 & 86822 & 173863.5 & 160 \\
DE-2LS-ISR012      & 87207 & 85771.5 & 172978.5 & 163 \\
DE-2LS-ISR020      & 86584 & 88195.5 & 174779.5 & 144 \\
\bottomrule
\end{tabular}}
\begin{flushleft}
\footnotesize Here, SR denotes the LS starting rate, BR denotes the LS budget rate, SG denotes the stagnation threshold, MC denotes the maximum number of LS calls, and ISR denotes the initial step rate.
\end{flushleft}
\end{table}

The ablation results show that DE-2LS achieves the highest
overall U-score among all tested variants. It obtains a total U-score of
178749, which is higher than the Base variant and all other local-search
configurations. The improvement comes from both accuracy and speed: DE-2LS
achieves the highest accuracy score of 88801.5 and the highest speed score of
89947.5. This suggests that reducing the local-search budget to 0.5\% of the
maximum evaluation budget is beneficial. The smaller budget preserves more
evaluations for the original RDEx population search while still allowing the
algorithm to perform useful late-stage refinement.

The results also show that a more aggressive local-search design is not always
beneficial. For example, DE-2LS-BR015 uses a larger local-search budget, but its
U-score is lower than that of DE-2LS. Similarly, DE-2LS-SR80 and DE-2LS-SR80SG12 start the local search earlier, but both variants obtain lower
overall U-scores. This indicates that premature local refinement may interrupt
the population-level exploration of RDEx. DE-2LS-SR90 obtains the same rank sum
as DE-2LS, but its U-score is lower. Therefore, the selected DE-2LS setting provides a better balance between late-stage exploitation and
the preservation of the original RDEx search dynamics.

The parameter settings of the finalized DE-2LS are summarized in Table \ref{tab:rdex2ls_general_params}. The initial RDEx front size is
$10D$, and the auxiliary population array has size $2\times 10D$. All local-search
evaluations are counted within the same global function-evaluation budget. The late local-search settings are as the final DE-2LS version uses a small local-search budget of 0.5\% of
\texttt{MaxFEval} per LS call and allows at most two LS calls. Therefore, the
maximum LS budget is about 1\% of the total evaluation budget. This setting keeps
local search as a controlled polishing component rather than allowing it to
dominate the evolutionary search.

\begin{table}[!h]
\centering
\caption{General parameter settings of DE-2LS.}
\label{tab:rdex2ls_general_params}
\renewcommand{\arraystretch}{1.15}
\resizebox{\columnwidth}{!}{
\begin{tabular}{lll}
\toprule
Parameter & Value & Description \\
\midrule
Initial front size & $10D$ & Initial active RDEx front size \\
Allocated population array & $2\times10D$ & Current and successful candidates \\
Minimum front size & 4 & Final reduced front size \\
Memory size & 5 & Adaptive $F$ and $CR$ memory size \\
Initial \texttt{MemoryCr} & 1 & Initial crossover-rate memory \\
Initial \texttt{MemoryF} & 0.3 & Initial scaling-factor memory \\
Initial success rate & 0.5 & Initial adaptive sampling rate \\
\texttt{EB\_rate} & 0.7 & Initial EB hybridization rate \\
\texttt{LS\_SR} & 0.85 & Allow LS after 85\% of \texttt{MaxFEval} \\
\texttt{LS\_FSR} & 0.95 & Allow very-late LS after 95\% \\
\texttt{LS\_BR} & 0.005 & 0.5\% of \texttt{MaxFEval} per LS call \\
\texttt{LS\_MC} & 2 & Maximum number of LS calls \\
\texttt{LS\_SG} & 8 & Stagnation threshold \\
\texttt{LS\_ISR} & 0.015 & Initial coordinate step ratio \\
\bottomrule
\end{tabular}}
\end{table}

\subsection{Head-to-Head Comparison with RDEx}
\label{subsec:comparison_rdex}

Since DE-2LS is developed directly from RDEx, the most important comparison is
the head-to-head comparison between these two algorithms. Table
\ref{tab:rdex2ls_vs_rdex_summary} summarizes their total accuracy, speed, and
U-score values.

\begin{table}[!h]
\centering
\caption{Head-to-head comparison between DE-2LS and RDEx.}
\label{tab:rdex2ls_vs_rdex_summary}
\renewcommand{\arraystretch}{1.15}
\begin{tabular}{lccc}
\toprule
Algorithm & Accuracy & Speed & U-score \\
\midrule
DE-2LS & \textbf{17663} & \textbf{17567.5} & \textbf{35230.5} \\
RDEx     & 16637          & 16732.5          & 33369.5 \\
\midrule
Relative gain & \textbf{+6.17\%} & \textbf{+4.99\%} & \textbf{+5.58\%} \\
\bottomrule
\end{tabular}
\end{table}

The proposed DE-2LS improves the total U-score from 33369.5 to 35230.5,
corresponding to a relative gain of 5.58\% over RDEx. The improvement is not only
due to better final accuracy but also due to improved speed score. Specifically,
DE-2LS improves the accuracy score by 6.17\% and the speed score by 4.99\%.
This confirms that the late local-search component does not simply improve final
polishing at the cost of search speed. Instead, the selected LS configuration
strengthens the final solution quality while preserving the fast search behavior
of RDEx.

The function-wise results (Table \ref{tab:function-wise}) further indicate that DE-2LS improves the U-score on
most functions. This behavior supports the main design principle of DE-2LS:
the original RDEx should remain the main global and feasibility-oriented search
engine, while local search should be activated only as a late-stage refinement
operator. Because the LS module is budget-aware and feasibility-aware, it can
improve the best solution without destabilizing the population or consuming an
excessive number of evaluations.

\begin{table}[!h]\centering
\caption{Function-wise pairwise U-score comparison between DE-2LS and RDEx.}\label{tab:function-wise}
\resizebox{1\linewidth}{!}{\begin{tabular}{c|cc|cc|cc}\hline\multirow{2}{*}{Func}& \multicolumn{2}{c|}{Accuracy}& \multicolumn{2}{c|}{Speed}& \multicolumn{2}{c}{U-score} \\\cline{2-7}& DE-2LS & RDEx& DE-2LS & RDEx& DE-2LS & RDEx \\\hline
F1	&		612.5	&		612.5	&		596	&	\cellcolor{gray!75}	629	&		1208.5	&	\cellcolor{gray!75}	1241.5	\\
F2	&		612.5	&		612.5	&		603.5	&	\cellcolor{gray!75}	621.5	&		1216	&	\cellcolor{gray!75}	1234	\\
F3	&		541	&	\cellcolor{gray!75}	684	&		541	&	\cellcolor{gray!75}	684	&		1082	&	\cellcolor{gray!75}	1368	\\
F4	&	\cellcolor{gray!75}	620.5	&		604.5	&		607.5	&	\cellcolor{gray!75}	617.5	&	\cellcolor{gray!75}	1228	&		1222	\\
F5	&		612.5	&		612.5	&	\cellcolor{gray!75}	704	&		521	&	\cellcolor{gray!75}	1316.5	&		1133.5	\\
F6	&	\cellcolor{gray!75}	650	&		575	&	\cellcolor{gray!75}	641.5	&		583.5	&	\cellcolor{gray!75}	1291.5	&		1158.5	\\
F7	&	\cellcolor{gray!75}	690	&		535	&		568.5	&	\cellcolor{gray!75}	656.5	&	\cellcolor{gray!75}	1258.5	&		1191.5	\\
F8	&		612.5	&		612.5	&		593	&	\cellcolor{gray!75}	632	&		1205.5	&	\cellcolor{gray!75}	1244.5	\\
F9	&		612.5	&		612.5	&		596	&	\cellcolor{gray!75}	629	&		1208.5	&	\cellcolor{gray!75}	1241.5	\\
F10	&		612.5	&		612.5	&	\cellcolor{gray!75}	661	&		564	&	\cellcolor{gray!75}	1273.5	&		1176.5	\\
F11	&	\cellcolor{gray!75}	622.5	&		602.5	&	\cellcolor{gray!75}	636.5	&		588.5	&	\cellcolor{gray!75}	1259	&		1191	\\
F12	&		610.5	&	\cellcolor{gray!75}	614.5	&	\cellcolor{gray!75}	639	&		586	&	\cellcolor{gray!75}	1249.5	&		1200.5	\\
F13	&		612.5	&		612.5	&	\cellcolor{gray!75}	674.5	&		550.5	&	\cellcolor{gray!75}	1287	&		1163	\\
F14	&	\cellcolor{gray!75}	637.5	&		587.5	&	\cellcolor{gray!75}	634	&		591	&	\cellcolor{gray!75}	1271.5	&		1178.5	\\
F15	&	\cellcolor{gray!75}	622	&		603	&		609	&	\cellcolor{gray!75}	616	&	\cellcolor{gray!75}	1231	&		1219	\\
F16	&	\cellcolor{gray!75}	648.5	&		576.5	&	\cellcolor{gray!75}	630	&		595	&	\cellcolor{gray!75}	1278.5	&		1171.5	\\
F17	&		612.5	&		612.5	&		586	&	\cellcolor{gray!75}	639	&		1198.5	&	\cellcolor{gray!75}	1251.5	\\
F18	&	\cellcolor{gray!75}	650	&		575	&	\cellcolor{gray!75}	641	&		584	&	\cellcolor{gray!75}	1291	&		1159	\\
F19	&		612.5	&		612.5	&	\cellcolor{gray!75}	682	&		543	&	\cellcolor{gray!75}	1294.5	&		1155.5	\\
F20	&	\cellcolor{gray!75}	651	&		574	&	\cellcolor{gray!75}	633	&		592	&	\cellcolor{gray!75}	1284	&		1166	\\
F21	&	\cellcolor{gray!75}	703.5	&		521.5	&	\cellcolor{gray!75}	656.5	&		568.5	&	\cellcolor{gray!75}	1360	&		1090	\\
F22	&		612.5	&		612.5	&	\cellcolor{gray!75}	642	&		583	&	\cellcolor{gray!75}	1254.5	&		1195.5	\\
F23	&		612.5	&		612.5	&	\cellcolor{gray!75}	636.5	&		588.5	&	\cellcolor{gray!75}	1249	&		1201	\\
F24	&		612.5	&		612.5	&		584	&	\cellcolor{gray!75}	641	&		1196.5	&	\cellcolor{gray!75}	1253.5	\\
F25	&	\cellcolor{gray!75}	646	&		579	&	\cellcolor{gray!75}	635.5	&		589.5	&	\cellcolor{gray!75}	1281.5	&		1168.5	\\
F26	&		612.5	&		612.5	&	\cellcolor{gray!75}	629	&		596	&	\cellcolor{gray!75}	1241.5	&		1208.5	\\
F27	&	\cellcolor{gray!75}	739	&		486	&	\cellcolor{gray!75}	644	&		581	&	\cellcolor{gray!75}	1383	&		1067	\\
F28	&	\cellcolor{gray!75}	668.5	&		556.5	&	\cellcolor{gray!75}	663	&		562	&	\cellcolor{gray!75}	1331.5	&		1118.5	\\\hline
\bf Total	&	\cellcolor{gray!75}	17663	&		16637	&	\cellcolor{gray!75}	17567.5	&		16732.5	&	\cellcolor{gray!75} 35230.5	&		33369.5	\\\hline
\end{tabular}}\end{table}

\subsection{Comparison with Competitive Algorithms}
\label{subsec:comparison_four_algorithms}
Table~\ref{tab:overall_four_algorithm_summary} summarizes the total accuracy,
speed, U-score, and rank values of DE-2LS, RDEx, CL-SRDE, and UDE-III over all
benchmark functions. DE-2LS obtains the highest overall U-score of 80968 and the best total rank of 48. Compared with the original RDEx, the U-score improves
from 78303.5 to 80968, corresponding to a relative gain of 3.40\%. This
indicates that the proposed late local-search component improves the overall
performance of RDEx under the combined speed-accuracy scoring criterion.

\begin{table}[!h]
\centering
\caption{Overall comparison of DE-2LS with competitive algorithms.}
\label{tab:overall_four_algorithm_summary}
\resizebox{\columnwidth}{!}{
\begin{tabular}{lrrrrc}
\toprule
Algorithm & Accuracy & Speed & U-score & Rank & Relative to RDEx \\
\midrule
DE-2LS & 35325 & \textbf{45643} & \textbf{80968} & \textbf{48} & \textbf{+3.40\%} \\
RDEx     & 33822 & 44481.5          & 78303.5          & 61          & baseline\\
UDE-III  & \textbf{38097} & 31123.5 & 69220.5 & 76 & $-11.60$\% \\
CL-SRDE  & 31356 & 17352          & 48708          & 95          & $-37.80$\% \\
\bottomrule
\end{tabular}}
\end{table}

\begin{figure}[!h]
    \centering
    \includegraphics[width=\linewidth]{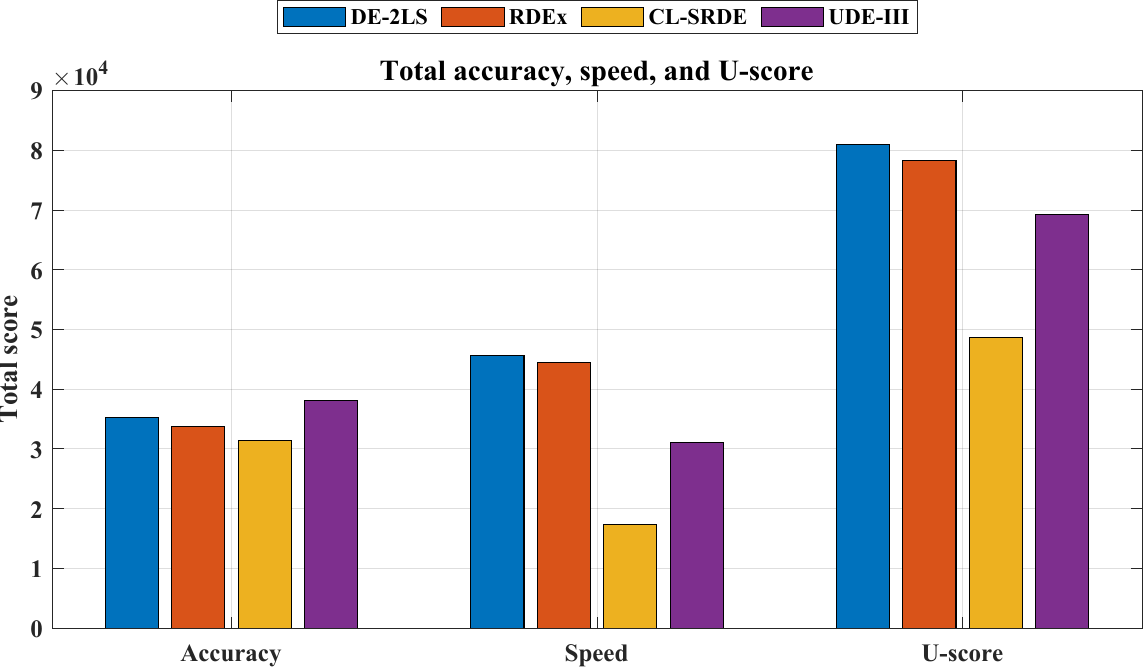}
   \caption{Total accuracy, speed, and U-score comparison of DE-2LS, RDEx, CL-SRDE, and UDE-III over 28 constrained benchmark functions.}\label{fig:total_accuracy_speed_uscore}
\end{figure}

A useful observation from Table~\ref{tab:overall_four_algorithm_summary} is that
UDE-III achieves the highest total accuracy score, whereas DE-2LS achieves the
highest speed score. However, the speed score of UDE-III is substantially lower
than that of DE-2LS and RDEx. Consequently, its final U-score remains lower
than both RDEx-based methods. This suggests that UDE-III can produce strong final
solutions on some functions, but it is less competitive when both accuracy and
speed are considered simultaneously. In contrast, DE-2LS provides a better
balance: it retains the fast convergence behavior of RDEx while improving its
final solution quality through a controlled late-stage local search. Fig.~\ref{fig:total_accuracy_speed_uscore} visually summarizes the aggregate accuracy, speed, and U-score values reported in Table~\ref{tab:overall_four_algorithm_summary}.

\begin{table*}[!h]\centering\caption{Function-wise comparison of DE-2LS, RDEx, CL-SRDE, and UDE-III in terms of accuracy, speed, U-score, and rank.}\label{tab:functionwise_four_algorithms}\scriptsize
\setlength{\tabcolsep}{2.5pt}\resizebox{0.93\textwidth}{!}{\begin{tabular}{c|llll|llll|llll|cccc}\hline\multirow{2}{*}{Func}& \multicolumn{4}{c|}{Accuracy}& \multicolumn{4}{c|}{Speed}& \multicolumn{4}{c|}{U-score}& \multicolumn{4}{c}{Rank} \\\cline{2-17}& DE-2LS & RDEx & CL-SRDE & UDE-III& DE-2LS & RDEx & CL-SRDE & UDE-III& DE-2LS & RDEx & CL-SRDE & UDE-III& DE-2LS & RDEx & CL-SRDE & UDE-III \\\hline
F1	&		1237.5	&		1237.5	&		1237.5	&		1237.5	&		1841	&	\cellcolor{gray!75}	1875.5	&	300	&		933.5	&		3078.5	&	\cellcolor{gray!75}	3113	&		1537.5	&		2171	&		2	&	\cellcolor{gray!75}	1	&		4	&		3	\\
F2	&		1237.5	&		1237.5	&		1237.5	&		1237.5	&		1853.5	&	\cellcolor{gray!75}	1871	&	300	&		925.5	&		3091	&	\cellcolor{gray!75}	3108.5	&		1537.5	&		2163	&		2	&	\cellcolor{gray!75}	1	&		4	&		3	\\
F3	&		1121	&		1309	&		368	&	\cellcolor{gray!75}	2152	&		1126	&		1310	&	362	&	\cellcolor{gray!75}	2152	&		2247	&		2619	&		730	&	\cellcolor{gray!75}	4304	&		3	&		2	&		4	&	\cellcolor{gray!75}	1	\\
F4	&		1240.5	&		1229.5	&		305	&	\cellcolor{gray!75}	2175	&		1227.5	&		1244.5	&	336	&	\cellcolor{gray!75}	2142	&		2468	&		2474	&		641	&	\cellcolor{gray!75}	4317	&		3	&		2	&		4	&	\cellcolor{gray!75}	1	\\
F5	&		1237.5	&		1237.5	&		1237.5	&		1237.5	&	\cellcolor{gray!75}	1684	&		1428.5	&	300	&		1537.5	&	\cellcolor{gray!75}	2921.5	&		2666	&		1537.5	&		2775	&	\cellcolor{gray!75}	1	&		3	&		4	&		2	\\
F6	&	\cellcolor{gray!75}	1275	&		1125	&	\cellcolor{gray!75}	1275	&	\cellcolor{gray!75}	1275	&	\cellcolor{gray!75}	1891.5	&		1754.5	&	339.5	&		964.5	&	\cellcolor{gray!75}	3166.5	&		2879.5	&		1614.5	&		2239.5	&	\cellcolor{gray!75}	1	&		2	&		4	&		3	\\
F7	&	\cellcolor{gray!75}	1720	&		1443	&		1100	&		687	&		1658.5	&	\cellcolor{gray!75}	1778.5	&	651	&		862	&	\cellcolor{gray!75}	3378.5	&		3221.5	&		1751	&		1549	&	\cellcolor{gray!75}	1	&		2	&		3	&		4	\\
F8	&		1237.5	&		1237.5	&		1237.5	&		1237.5	&		1843	&	\cellcolor{gray!75}	1882	&	925	&		300	&		3080.5	&	\cellcolor{gray!75}	3119.5	&		2162.5	&		1537.5	&		2	&	\cellcolor{gray!75}	1	&		3	&		4	\\
F9	&		1237.5	&		1237.5	&		1237.5	&		1237.5	&		1817	&	\cellcolor{gray!75}	1837	&	812.5	&		483.5	&		3054.5	&	\cellcolor{gray!75}	3074.5	&		2050	&		1721	&		2	&	\cellcolor{gray!75}	1	&		3	&		4	\\
F10	&		1237.5	&		1237.5	&		1237.5	&		1237.5	&	\cellcolor{gray!75}	1911	&		1814	&	925	&		300	&	\cellcolor{gray!75}	3148.5	&		3051.5	&		2162.5	&		1537.5	&	\cellcolor{gray!75}	1	&		2	&		3	&		4	\\
F11	&	\cellcolor{gray!75}	1650.5	&		1635.5	&		1259	&		405	&	\cellcolor{gray!75}	1459	&		1375.5	&	1375.5	&		740	&	\cellcolor{gray!75}	3109.5	&		3011	&		2634.5	&		1145	&	\cellcolor{gray!75}	1	&		2	&		3	&		4	\\
F12	&		652.5	&		702.5	&		1566	&	\cellcolor{gray!75}	2029	&		1086	&		1084.5	&	729.5	&	\cellcolor{gray!75}	2050	&		1738.5	&		1787	&		2295.5	&	\cellcolor{gray!75}	4079	&		4	&		3	&		2	&	\cellcolor{gray!75}	1	\\
F13	&	\cellcolor{gray!75}	1275	&	\cellcolor{gray!75}	1275	&	\cellcolor{gray!75}	1275	&		1125	&	\cellcolor{gray!75}	1749	&		1620	&	350	&		1231	&	\cellcolor{gray!75}	3024	&		2895	&		1625	&		2356	&	\cellcolor{gray!75}	1	&		2	&		4	&		3	\\
F14	&		987.5	&		937.5	&		850	&	\cellcolor{gray!75}	2175	&	\cellcolor{gray!75}	1882	&		1786	&	601	&		681	&	\cellcolor{gray!75}	2869.5	&		2723.5	&		1451	&		2856	&	\cellcolor{gray!75}	1	&		3	&		4	&		2	\\
F15	&		944	&		917	&		914	&	\cellcolor{gray!75}	2175	&		1132.5	&		1149	&	562	&	\cellcolor{gray!75}	2106.5	&		2076.5	&		2066	&		1476	&	\cellcolor{gray!75}	4281.5	&		2	&		3	&		4	&	\cellcolor{gray!75}	1	\\
F16	&		1150.5	&		1030.5	&		594	&	\cellcolor{gray!75}	2175	&		1233	&		1186	&	356	&	\cellcolor{gray!75}	2175	&		2383.5	&		2216.5	&		950	&	\cellcolor{gray!75}	4350	&		2	&		3	&		4	&	\cellcolor{gray!75}	1	\\
F17	&		975	&		975	&		975	&	\cellcolor{gray!75}	2025	&		1836	&	\cellcolor{gray!75}	1889	&	633	&		592	&		2811	&	\cellcolor{gray!75}	2864	&		1608	&		2617	&		2	&	\cellcolor{gray!75}	1	&		4	&		3	\\
F18	&		1800	&		1662	&		1188	&		300	&	\cellcolor{gray!75}	1890	&		1808	&	788	&		464	&	\cellcolor{gray!75}	3690	&		3470	&		1976	&		764	&	\cellcolor{gray!75}	1	&		2	&		3	&		4	\\
F19	&	\cellcolor{gray!75}	1862.5	&		1862.5	&		300	&		925	&		1307	&		1168	&	300	&	\cellcolor{gray!75}	2175	&	\cellcolor{gray!75}	3169.5	&		3030.5	&		600	&		3100	&	\cellcolor{gray!75}	1	&		3	&		4	&		2	\\
F20	&	\cellcolor{gray!75}	1840	&		1741	&		427	&		942	&	\cellcolor{gray!75}	1837	&		1793	&	445	&		875	&	\cellcolor{gray!75}	3677	&		3534	&		872	&		1817	&	\cellcolor{gray!75}	1	&		2	&		4	&		3	\\
F21	&		1333.5	&		1049.5	&		685	&	\cellcolor{gray!75}	1882	&		1308.5	&		1106.5	&	653	&	\cellcolor{gray!75}	1882	&		2642	&		2156	&		1338	&	\cellcolor{gray!75}	3764	&		2	&		3	&		4	&	\cellcolor{gray!75}	1	\\
F22	&	\cellcolor{gray!75}	1362.5	&		1362.5	&		1362.5	&		862.5	&	\cellcolor{gray!75}	1892	&		1833	&	925	&		300	&	\cellcolor{gray!75}	3254.5	&		3195.5	&		2287.5	&		1162.5	&	\cellcolor{gray!75}	1	&		2	&		3	&		4	\\
F23	&		912.5	&		912.5	&	\cellcolor{gray!75}	2175	&		950	&	\cellcolor{gray!75}	1886.5	&		1838.5	&	925	&		300	&		2799	&		2751	&	\cellcolor{gray!75}	3100	&		1250	&		2	&		3	&	\cellcolor{gray!75}	1	&		4	\\
F24	&		612.5	&		612.5	&		1550	&	\cellcolor{gray!75}	2175	&		1179.5	&		1273.5	&	485.5	&	\cellcolor{gray!75}	2011.5	&		1792	&		1886	&		2035.5	&	\cellcolor{gray!75}	4186.5	&		4	&		3	&		2	&	\cellcolor{gray!75}	1	\\
F25	&		1099	&		976	&		700	&	\cellcolor{gray!75}	2175	&		1501.5	&		1410.5	&	369	&	\cellcolor{gray!75}	1669	&		2600.5	&		2386.5	&		1069	&	\cellcolor{gray!75}	3844	&		2	&		3	&		4	&	\cellcolor{gray!75}	1	\\
F26	&		1162.5	&		1162.5	&		1162.5	&	\cellcolor{gray!75}	1462.5	&	\cellcolor{gray!75}	1878	&		1846	&	802.5	&		423.5	&	\cellcolor{gray!75}	3040.5	&		3008.5	&		1965	&		1886	&	\cellcolor{gray!75}	1	&		2	&		3	&		4	\\
F27	&		1405	&		1120	&	\cellcolor{gray!75}	2125	&		300	&	\cellcolor{gray!75}	1844.5	&		1782	&	776	&		547.5	&	\cellcolor{gray!75}	3249.5	&		2902	&		2901	&		847.5	&	\cellcolor{gray!75}	1	&		2	&		3	&		4	\\
F28	&		1518.5	&		1356.5	&	\cellcolor{gray!75}	1775	&		300	&	\cellcolor{gray!75}	1888	&		1737	&	1025	&		300	&	\cellcolor{gray!75}	3406.5	&		3093.5	&		2800	&		600	&	\cellcolor{gray!75}	1	&		2	&		3	&		4	\\\hline
\bf Total	&		35325	&		33822	&		31356	&	\cellcolor{gray!75}	38097	&	\cellcolor{gray!75}	45643	&		44481.5	&	17352	&		31123.5	&	\cellcolor{gray!75}	80968	&		78303.5	&		48708	&		69220.5	&	\cellcolor{gray!75}	48	&		61	&		95	&		76	\\\hline
\end{tabular}}\end{table*}
\begin{table*}[!h]
\centering
\caption{Mean and standard deviation comparison of DE-2LS, RDEx, CL-SRDE, and UDE-III.}\label{tab:mean_std_comparison}
\scriptsize
\setlength{\tabcolsep}{2.5pt}\resizebox{0.93\textwidth}{!}{\begin{tabular}{c|c|c|c|c}\hline
\multirow{2}{*}{Func}& \multicolumn{1}{c}{DE-2LS}& \multicolumn{1}{c}{RDEx}& \multicolumn{1}{c}{CL-SRDE}& \multicolumn{1}{c}{UDE-III} \\
\cline{2-5}& Mean $\pm$ STD& Mean $\pm$ STD& Mean $\pm$ STD& Mean $\pm$ STD \\\hline
F1	&	\cellcolor{gray!75}	0.000E+00	$\pm$	0.000E+00	&	\cellcolor{gray!75}	0.000E+00	$\pm$	0.000E+00	&	\cellcolor{gray!75}	0.000E+00	$\pm$	0.000E+00	&	\cellcolor{gray!75}	0.000E+00	$\pm$	0.000E+00	\\
F2	&	\cellcolor{gray!75}	0.000E+00	$\pm$	0.000E+00	&	\cellcolor{gray!75}	0.000E+00	$\pm$	0.000E+00	&	\cellcolor{gray!75}	0.000E+00	$\pm$	0.000E+00	&	\cellcolor{gray!75}	0.000E+00	$\pm$	0.000E+00	\\
F3	&		3.063E+02	$\pm$	1.030E+02	&		2.655E+02	$\pm$	6.341E+01	&		5.021E+02	$\pm$	1.074E+02	&	\cellcolor{gray!75}	9.221E+01	$\pm$	5.651E+01	\\
F4	&		1.679E+01	$\pm$	4.125E+00	&		1.633E+01	$\pm$	2.417E+00	&		4.625E+01	$\pm$	8.876E+00	&	\cellcolor{gray!75}	6.515E+00	$\pm$	6.921E+00	\\
F5	&	\cellcolor{gray!75}	0.000E+00	$\pm$	0.000E+00	&	\cellcolor{gray!75}	0.000E+00	$\pm$	0.000E+00	&	\cellcolor{gray!75}	0.000E+00	$\pm$	0.000E+00	&	\cellcolor{gray!75}	0.000E+00	$\pm$	0.000E+00	\\
F6	&	\cellcolor{gray!75}	0.000E+00		0.000E+00	&		96\%			&	\cellcolor{gray!75}	0.000E+00	$\pm$	0.000E+00	&	\cellcolor{gray!75}	0.000E+00	$\pm$	0.000E+00	\\
F7	&	\cellcolor{gray!75}	-1.136E+03	$\pm$	4.103E+02	&		-1.008E+03	$\pm$	5.117E+02	&		-8.111E+02	$\pm$	2.045E+02	&		-6.732E+02	$\pm$	1.517E+02	\\
F8	&	\cellcolor{gray!75}	-2.840E-04	$\pm$	0.000E+00	&	\cellcolor{gray!75}	-2.840E-04	$\pm$	0.000E+00	&	\cellcolor{gray!75}	-2.840E-04	$\pm$	2.769E-16	&	\cellcolor{gray!75}	-2.840E-04	$\pm$	5.972E-12	\\
F9	&	\cellcolor{gray!75}	-2.666E-03	$\pm$	1.328E-18	&	\cellcolor{gray!75}	-2.666E-03	$\pm$	1.328E-18	&	\cellcolor{gray!75}	-2.666E-03	$\pm$	1.328E-18	&	\cellcolor{gray!75}	-2.666E-03	$\pm$	8.852E-19	\\
F10	&	\cellcolor{gray!75}	-1.028E-04	$\pm$	0.000E+00	&	\cellcolor{gray!75}	-1.028E-04	$\pm$	0.000E+00	&	\cellcolor{gray!75}	-1.028E-04	$\pm$	2.000E-16	&	\cellcolor{gray!75}	-1.028E-04	$\pm$	1.383E-20	\\
F11	&	\cellcolor{gray!75}	-5.113E+00	$\pm$	6.011E+00	&		-4.833E+00	$\pm$	4.952E+00	&		-8.303E-01	$\pm$	1.606E-01	&		40\%			\\
F12	&		9.775E+00	$\pm$	5.256E-04	&		9.312E+00	$\pm$	1.604E+00	&		8.848E+00	$\pm$	2.167E+00	&	\cellcolor{gray!75}	3.992E+00	$\pm$	2.225E-02	\\
F13	&	\cellcolor{gray!75}	0.000E+00	$\pm$	0.000E+00	&	\cellcolor{gray!75}	0.000E+00	$\pm$	0.000E+00	&	\cellcolor{gray!75}	0.000E+00	$\pm$	0.000E+00	&		4.784E-01	$\pm$	1.322E+00	\\
F14	&		1.412E+00	$\pm$	1.738E-02	&		1.419E+00	$\pm$	2.883E-02	&		1.457E+00	$\pm$	4.404E-02	&	\cellcolor{gray!75}	1.409E+00	$\pm$	6.799E-16	\\
F15	&		5.875E+00	$\pm$	1.042E+00	&		5.875E+00	$\pm$	1.042E+00	&		6.503E+00	$\pm$	1.749E+00	&	\cellcolor{gray!75}	2.356E+00	$\pm$	1.441E-06	\\
F16	&		2.111E+01	$\pm$	3.272E+00	&		2.187E+01	$\pm$	3.846E+00	&		2.256E+01	$\pm$	2.791E+00	&	\cellcolor{gray!75}	0.000E+00	$\pm$	0.000E+00	\\
F17	&		0\% (31)			&		0\% (31)			&		0\% (31)			&	\cellcolor{gray!75}	0\% (29.42)			\\
F18	&	\cellcolor{gray!75}	3.652E+01	$\pm$	1.450E-14	&	\cellcolor{gray!75}	3.652E+01	$\pm$	3.317E-05	&	\cellcolor{gray!75}	3.652E+01	$\pm$	6.325E-06	&		3.655E+01	$\pm$	1.394E-01	\\
F19	&	\cellcolor{gray!75}	0\% (42734.8)	$\pm$		&	\cellcolor{gray!75}	0\% (42734.8)			&		0\% (42749.8)	$\pm$		&		0\% (42749.8)			\\
F20	&	\cellcolor{gray!75}	1.351E+00	$\pm$	2.036E-01	&		1.394E+00	$\pm$	1.949E-01	&		2.788E+00	$\pm$	6.979E-01	&		1.851E+00	$\pm$	2.882E-01	\\
F21	&		2.239E+01	$\pm$	8.832E+00	&		2.928E+01	$\pm$	1.025E+01	&		2.598E+01	$\pm$	1.026E+01	&	\cellcolor{gray!75}	9.280E+00	$\pm$	8.507E+00	\\
F22	&	\cellcolor{gray!75}	0.000E+00	$\pm$	0.000E+00	&	\cellcolor{gray!75}	0.000E+00	$\pm$	0.000E+00	&	\cellcolor{gray!75}	0.000E+00	$\pm$	0.000E+00	&		2.556E+01	$\pm$	4.944E+01	\\
F23	&	\cellcolor{gray!75}	1.409E+00	$\pm$	2.266E-16	&	\cellcolor{gray!75}	1.409E+00	$\pm$	2.266E-16	&	\cellcolor{gray!75}	1.409E+00	$\pm$	6.799E-16	&		1.450E+00	$\pm$	4.432E-02	\\
F24	&		5.498E+00	$\pm$	9.065E-16	&		5.498E+00	$\pm$	9.065E-16	&		5.498E+00	$\pm$	9.065E-16	&	\cellcolor{gray!75}	2.356E+00	$\pm$	1.000E-07	\\
F25	&		2.400E+01	$\pm$	4.680E+00	&		2.482E+01	$\pm$	4.988E+00	&		2.450E+01	$\pm$	3.483E+00	&	\cellcolor{gray!75}	2.513E-01	$\pm$	1.257E+00	\\
F26	&		0\% (31)			&		0\% (31)			&		0\% (31)			&	\cellcolor{gray!75}	0\% (30.55)			\\
F27	&	\cellcolor{gray!75}	3.652E+01	$\pm$	6.272E-05	&	\cellcolor{gray!75}	3.652E+01	$\pm$	1.166E-04	&	\cellcolor{gray!75}	3.652E+01	$\pm$	2.618E-05	&		3.662E+01	$\pm$	3.614E-01	\\
F28	&		0\% (42750.9)			&		0\% (42753.8)			&	\cellcolor{gray!75}	0\% (42749.8)			&		0\% (42881.8)			\\\hline
\end{tabular}}\end{table*}
Table~\ref{tab:functionwise_four_algorithms} provides the function-wise comparison among the four algorithms. The results show that DE-2LS obtains
rank 1 on many functions and achieves the best total rank across the full test
suite. The function-wise results also show that the advantage of DE-2LS is not
limited to a small number of functions. Instead, the proposed method improves the
combined U-score on a broad set of problems, especially those where additional
late-stage refinement can improve the final incumbent without delaying the main
search process. RDEx remains highly competitive and wins on several functions,
which confirms that the base RDEx framework is already strong. Nevertheless, the
overall rank and U-score indicate that adding a lightweight late local search
improves its robustness across the benchmark set.

The same table also reveals the complementary behavior of the compared
algorithms. UDE-III performs particularly well on some functions in terms of
accuracy, but its lower speed score reduces its final U-score. CL-SRDE is
competitive on a few individual cases but obtains the lowest total U-score and
largest rank sum among the four algorithms. These results support the main
motivation of DE-2LS: instead of replacing RDEx's evolutionary search engine,
a small and carefully triggered local-search phase can enhance exploitation while
preserving the original speed advantage.

Table~\ref{tab:mean_std_comparison} reports the mean and standard deviation of
the final objective/error values. This table gives a complementary view to the
U-score-based comparison. While the U-score evaluates both speed and accuracy
through pairwise comparisons, the mean--standard deviation results directly show
the final solution quality and stability of each algorithm. The results indicate
that no single algorithm dominates all functions in terms of final mean value.
For example, UDE-III achieves the best final mean on several functions, which is
consistent with its high total accuracy score. However, DE-2LS remains highly
competitive in final accuracy and obtains the best or tied-best mean value on
several functions.

The mean--standard deviation results further clarify why DE-2LS achieves the
best overall U-score. Its advantage does not come from being the most accurate
method for every function. Rather, DE-2LS combines competitive final accuracy
with consistently stronger speed. The low or comparable standard deviation on
several functions also suggests that the late local-search module does not
destabilize RDEx. Instead, it acts as a controlled polishing mechanism that can
improve the incumbent solution while keeping the stochastic behavior of the
original algorithm stable.

Overall, the three tables lead to a consistent conclusion. DE-2LS improves the
original RDEx by strengthening late-stage exploitation, but it does so without
sacrificing the speed-oriented behavior that makes RDEx competitive under the
U-score criterion. Although UDE-III is strong in final accuracy on some
functions, DE-2LS achieves the best overall trade-off between accuracy, speed,
and rank. Therefore, the proposed DE-2LS provides a more balanced and robust
performance profile for the considered constrained optimization benchmark.
\section{Conclusion}
\label{sec:conclusion}

This paper proposes DE-2LS, a late-stage local-search-enhanced version of RDEx for constrained single-objective numerical optimization. The proposed method preserves the original RDEx framework and adds a lightweight coordinate-pattern local search around the current best solution. 
The ablation results show that the finalized DE-2LS setting provides the best balance between exploitation and preservation of the original RDEx search behavior. It achieved the highest ablation U-score of 178749, outperforming variants with earlier LS activation, larger LS budget, different numbers of LS calls, and different step-size settings.
The direct comparison with RDEx shows that DE-2LS improves the U-score from 33369.5 to 35230.5, corresponding to a 5.58\% gain. In the four-algorithm comparison, DE-2LS achieved the highest overall U-score of 80968 and the best total rank of 48. These results confirm that the proposed late LS module improves RDEx without weakening its speed advantage.
Future work will investigate adaptive control of the LS budget, step size, and activation condition, and will extend the evaluation to additional dimensions and constrained benchmark suites.

\section*{Acknowledgment}
The authors sincerely thank Prof. P. N. Suganthan and the organizers of the IEEE CEC numerical optimization competition for providing the official U-score evaluation procedure and related competition resources. These materials were highly valuable for implementing the evaluation framework and ensuring consistency with the intended competition-based assessment protocol used in this work. The authors also thank the authors of RDEx for providing the source code, which facilitated the implementation of the baseline method and the development of the proposed DE-2LS algorithm.

\bibliographystyle{IEEEtran}
\bibliography{rdex_ls_cops}

\end{document}